\documentclass[runningheads]{llncs}

\usepackage{eccv}

\usepackage{eccvabbrv}
\usepackage{CJKutf8}
\usepackage{graphicx}
\usepackage{booktabs}
\usepackage[linesnumbered,ruled]{algorithm2e}
\usepackage[accsupp]{axessibility}  

\usepackage{hyperref}

\usepackage{orcidlink}

\begin{document}

\begin{CJK}{UTF8}{gbsn}
\title{Binomial Self-compensation for Motion Error in Dynamic 3D Scanning } 

\titlerunning{Binomial Self-compensation for Motion Error in Dynamic 3D Scanning}

\author{Geyou Zhang\inst{1}\orcidlink{0000-0002-3896-3238} \and
Ce Zhu\inst{1,*}\orcidlink{0000-0001-7607-707X}\and
Kai Liu\inst{2}\orcidlink{0000-0002-6433-6529}}
\authorrunning{G. Zhang et al.}

\institute{University of Electronic Science and Technology of China 
\\ \email{geyouzhang@foxmail.com, eczhu@uestc.edu.cn} \and
Sichuan University \\
\email{kailiu@scu.edu.cn}
}
\maketitle

\begin{abstract}
Phase shifting profilometry (PSP) is favored in high-precision 3D scanning due to its high accuracy, robustness, and pixel-wise property. However, a fundamental assumption of PSP that the object should remain static is violated in dynamic measurement, making PSP susceptible to object moving, resulting in ripple-like errors in the point clouds. We propose a pixel-wise and frame-wise loopable binomial self-compensation (BSC) algorithm to effectively and flexibly eliminate motion error in the four-step PSP. Our mathematical model demonstrates that by summing successive motion-affected phase frames weighted by binomial coefficients, motion error exponentially diminishes as the binomial order increases, accomplishing automatic error compensation through the motion-affected phase sequence, without the assistance of any intermediate variable. Extensive experiments show that our BSC outperforms the existing methods in reducing motion error, while achieving a depth map frame rate equal to the camera's acquisition rate (90 fps), enabling high-accuracy 3D reconstruction with a quasi-single-shot frame rate. The code is available
at \url{https://github.com/GeyouZhang/BSC}.

\end{abstract}

\section{Introduction}
\label{sec:intro}
In recent years, structured light (SL) 3D imaging has been penetrated into diverse fields, e.g., industrial inspection~\cite{qian2021high} and design~\cite{zhu2023high}, medical 3D reconstruction~\cite{logozzo2014recent, hirose2020200, geng2011structured, lv2023modeling}, virtual reality~\cite{ye2020accurate}, and digitalization of cultural heritage~\cite{honrado2004update}. Phase shifting profilometry (PSP)~\cite{srinivasan1984automated}, Fourier transform profilometry, binary code~\cite{posdamer1982surface}, and Gray code~\cite{sansoni1997three} are famous SL technologies that are widely applied in industry and scientific researches. Among the SL technologies above, PSP is especially favored in high-precision 3D scanning due to its high accuracy, robustness, and pixel-wise property~\cite{zuo2018phase}. Many studies have excavated the huge potentiality of PSP in various challenging scenes, such as low signal-to-noise~\cite{gupta2013structured,gupta2018geometric}, multi-path~\cite{zhang2019causes,zhang2021sparse}, high reflectivity~\cite{nayar2012diffuse}, and global illumination~\cite{chen2008modulated,gupta2012combined,achar2014multi}, et. al. 

However, a fundamental assumption of PSP that the object should remain static is violated in dynamic measurement, making PSP susceptible to object moving, resulting in ripple-like errors in the point clouds. We observed that the solutions of PSP in dynamic scenes are overlooked within the computer vision community. Concurrently, there are some existing methods in the optics metrology community to address this challenge, most of them can be categorized into three types: 1) motion-induced phase shifting estimation, 2) object tracking, and 3) automatic compensation. 

\textbf{Motion-induced phase shifting estimation}: In traditional PSP, the phase shifting of each pattern is equally distributed or pre-known values within one period of fringe, which becomes disorderly due to object motion. Thus, researchers estimate the motion-induced phase shifting according to the implicit motion information in the motion-affected phase frame~\cite{weise2007fast} or frames~\cite{liu2018motion,qian2019motion, liu2019real,wang2019motion}, thereby restoring the unknown motion-affected phase shifting value of each pattern, and finally solve the corrected phase by least squared method~\cite{liu20123d}. However, these methods may not effectively compensate for errors when the motion information can't be precisely predicted.

\textbf{Object tracking}: 
By tracking the position of the object by placing markers\cite{lu2013new} or using image feature matching algorithm\cite{lu2018general}, the motion error can be corrected by identifying the trajectory of the object~\cite{lu2013new,lu2017automated,lu2018general}. However, a limitation of the object tracking based methods is that they are tailored for 2D motion perpendicular to the sight line. Moreover, the accuracy and reliability of feature tracking are crucial issues when faced with textureless objects.

\textbf{Automatic compensation}: 
In automatic compensation methods, instead of computing motion-induced phase shifting value or tracking the object, an appropriate operator is selected to conduct on the whole captured image~\cite{wang2018motion,wu2022suppressing} or phase frame~\cite{guo2021real}, and further directly obtain the motion-error-free phase, in which the motion error will be automatically compensated.

Previous work on motion compensation typically estimates or generates certain intermediate variables, then calculates the motion-error-free phase based on these variables, either directly or indirectly. For example, in phase shifting estimation methods, the intermediate variables are motion-induced phase shifting values~\cite{liu2018motion,qian2019motion, liu2019real,wang2019motion}; in object tracking methods, the intermediate variable is the rigid transform matrix~\cite{lu2013new,lu2017automated,lu2018general}; in automatic compensation methods, the intermediate variables are the captured images after undergoing Hilbert transform~\cite{wang2018motion} or differential operator~\cite{wu2022suppressing}. Therefore, two questions naturally arise: \textbf{Why} not utilize the motion-affected phase itself to compensate for motion error without depending on any intermediate variables? Would this approach be more effective and flexible?

In this paper, without depending on any extra intermediate variable, we develop a binomial self-compensation (BSC) method that only utilizes the motion-affected phase sequence itself, thereby effectively and flexibly eliminating the motion error in four-step PSP. To achieve this, first, we design a paraxial binocular structured light system for reconstructing 3D point clouds using a single set of high-frequency fringe patterns. This system intentionally sets a very short baseline between two cameras to achieve a narrow disparity range, facilitating the use of high frequency fringes, while ensuring the baseline between the main camera and projector is long enough for adequate accuracy. Subsequently, we cyclically~\cite{zhong2013fast} project high-frequency patterns with $\pi/2$ phase shifting, and directly compute a sequence of phase frames from each three or four successive captured images. Finally, by adding up a sequence of successive phase frames weighted by binomial coefficients, we achieve self-compensation of motion errors. Our mathematical model demonstrates that motion error exponentially diminishes as the binomial order~$K$ increases.

Our BSC not only inherits the pixel-wise advantages of PSP, but also is of high temporal resolution to achieve quasi-single-shot 3D imaging frame rate because it's frame-wise loopable. Furthermore, our method offers flexibility and ease of deployment, acting as an enhanced, motion-error-free version of the four-step PSP, thus providing a plug-and-play enhancement for dynamic 3D scanning scenarios.

\section{Related Work}
In the introduction section, based on specific implementation details, we categorize the motion error compensation methods into three classes. From another perspective, we can also distinguish them based on their 3D reconstruction performance on temporal and spatial resolution. We examine the motion compensation methods from two perspectives: property of pixel-wise and frame-wise loopable. Pixel-wise methods compute phase and point clouds pixel by pixel, achieving spatial independence through this detailed granularity. Frame-wise loopable methods accomplish the same depth map frame rate as the camera acquisition frame rate, enabling a quasi-single-shot 3D reconstruction frame rate.

\textbf{Pixel-wise}: Several motion error compensation methods preserve the pixel-wise precision of traditional PSP, which utilizing two sampling phase frames to iteratively compute motion phase shifting~\cite{liu2018motion, wang2019motion}, automatically compensate for errors~\cite{guo2021real}, tracking the position of the object by placing markers\cite{lu2013new}, using image feature matching algorithm\cite{lu2018general} and correct motion error by identifying the trajectory of the object~\cite{lu2013new,lu2017automated,lu2018general}. These approaches build on the granularity of pixel-level adjustments. In contrast, the non-pixel-wise methods  conduct particular operator on local~\cite{wu2022suppressing,liu2019real,weise2007fast} or global image~\cite{zuo2018micro,qian2019motion,wang2018motion}, e.g., Fourier transform~\cite{zuo2018micro,qian2019motion}, Hilbert transform~\cite{wang2018motion}, differential operator~\cite{wu2022suppressing}, mean filter~\cite{liu2019real}, optimization algorithm~\cite{weise2007fast}, from which estimate the extra phase shifting induced by motion and compensate error~\cite{qian2019motion,liu2019real} or directly correct motion error~\cite{zuo2018micro,wu2022suppressing,wang2018motion,weise2007fast}. However, the pixel-wise merit of PSP is not preserved in these methods, theoretically diminishing their robustness in handling depth discontinuities. Concurrently, introducing extra non-pixel-wise operators also escalates the computational time required. 

\textbf{Frame-wise loopable}: By projecting high-frequency fringes and reconstructing 3D point clouds using a cyclic strategy~\cite{zhong2013fast}, some motion error compensation methods~\cite{weise2007fast,wang2018motion,liu2019real,guo2021real,wu2022suppressing} accomplish 3D reconstruction with quasi-single-shot frame rate, thus called frame-wise loopable methods in this paper. The cyclic strategy is not suitable for all motion compensation methods, some methods need to project an indivisible group of patterns~\cite{zuo2018micro,wang2019motion,qian2019motion}, such that the depth map frame rate is reduced severalfold.


\section{System Setup and Frequency Selection}
\begin{figure*}[t!]
    \centering
    \includegraphics[width=0.98\linewidth]{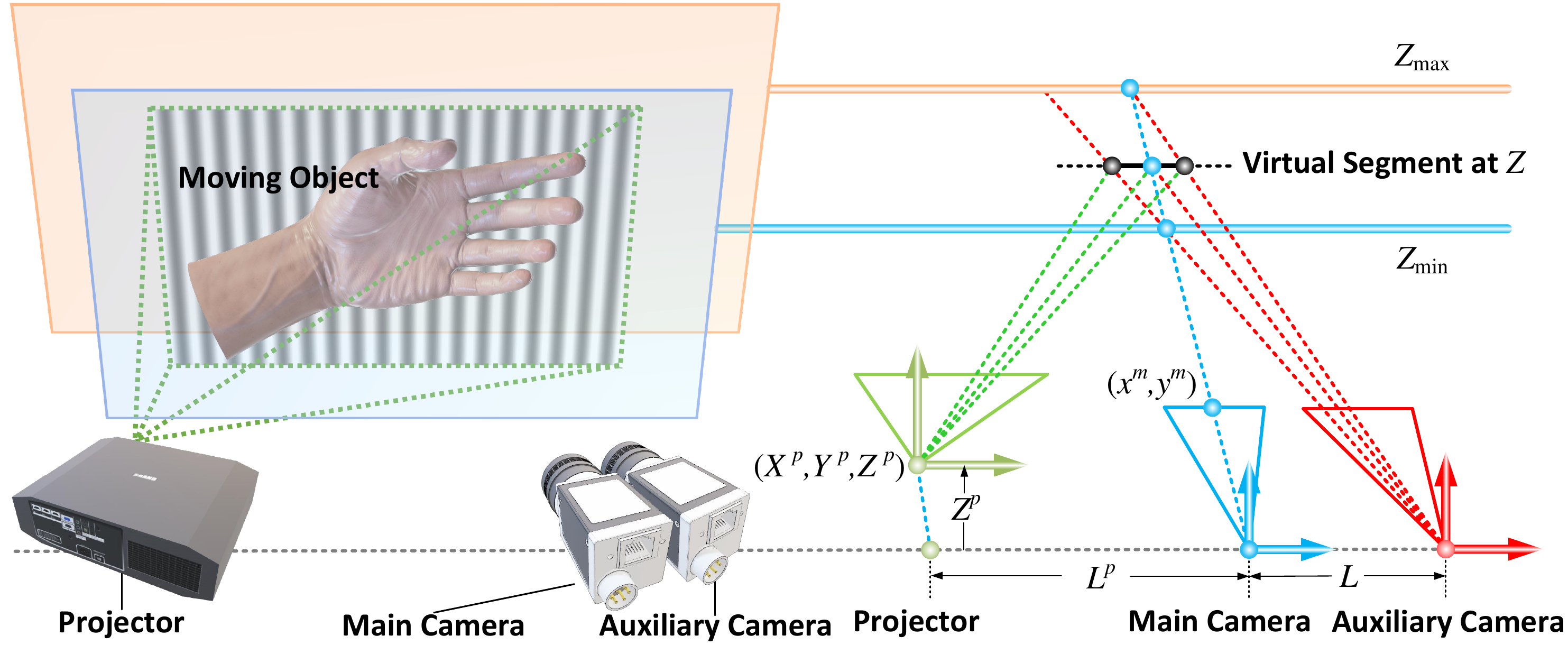}
    \caption{Diagram of our paraxial binocular structured light system.}
    \label{Fig:FigSystemDiagram}
\end{figure*}
Our paraxial binocular structured light system comprises a main camera, an auxiliary camera, and a projector, as depicted in Fig.~\ref{Fig:FigSystemDiagram}. Together, the main and auxiliary cameras form a paraxial setup to provide reference depth for phase unwrapping. Simultaneously, the combination of the main camera and projector facilitates precise 3D reconstruction. Following system calibration, we perform stereo rectification on the captured images to align the epipolar lines in the auxiliary system parallelly.

To ensure the simplicity of equations and convey intuitive geometrical relations, the following analysis in this section is under the assumption that the optical centers of the projector, main camera, and auxiliary camera are coplanar. Note that we build the world coordinate system to coincide with the main camera coordinate system. In reality, the projector may not be strictly coplanar to the other two, but the conclusions of our analysis are still effective.

The primary benefit of the binocular structured light (SL) system lies in its ability to perform stereo phase unwrapping (SPU) through geometric constraints, substantially decreasing the needed number of fringe patterns. To avoid phase unwrapping ambiguities, it should be ensured that only a few candidate points are within the depth range. As a result, traditional SPU approaches typically necessitate a short baseline between the main camera and the projector, which unfortunately compromises the accuracy of point clouds.

To address this, in this paper, we shift the constraint from the camera-projector baseline to the camera-camera baseline, thus allowing flexibility in the camera-projector distance and enhancing point cloud precision. This camera-camera baseline constraint is called the uniqueness constraint, ensuring that the phase remains single-valued within the feasible stereo matching range on the auxiliary camera, from which guarantees the uniqueness of the matching point candidate.

Given that stereo rectification has been applied to the cameras, we can disregard the Y axis. Assuming the 3D surface near the measured point is continuous and orthogonal to the Z axis, determining the depth range allows us to calculate the maximum fringe frequency. Consider a virtual segment, represented by a black dashed line in Fig.~\ref{Fig:FigSystemDiagram}, cut by the auxiliary camera's feasible region cone. The length of this virtual segment (black solid line in Fig.~\ref{Fig:FigSystemDiagram}) can be calculated as:
\begin{equation}
    {d} = \frac{{\left( {{Z_{\max }} - {Z_{\min }}} \right)ZL}}{{{Z_{\max }}{Z_{\min }}}}.
\end{equation}
As long as the projection of the virtual segment onto the projector space is distributed within one fringe period, the uniqueness constraint is satisfied, i.e.,
\begin{equation}
    \frac{{f_x^p{d}}}{{Z - {Z^p}}} < \frac{{{W^p}}}{f},
\end{equation}
where $f_x^p$ is the focal length of the projector, $W^p$ is the resolution of the projector along X axis, $f$ is the frequency of fringe pattern.

Finally, according to uniqueness constraint, we can obtain the limiting frequency of fringe as
\begin{equation}\label{EQ:UniquenessConstraint}
    f<\left( {1 - \frac{{{Z^p}}}{Z}} \right)\frac{{{Z_{\min }}{Z_{\max }}{W^p}}}{{\left( {{Z_{\max }} - {Z_{\min }}} \right)Lf_x^p}}.
\end{equation}
Actually, the uniqueness constraint is derived from an aggressive postulate that the 3D surface around the measured point is continuous and orthogonal with the Z axis, so the uniqueness constraint is a weak constraint. Thus, the camera-camera baseline should be as short as possible to ensure the uniqueness of phase value in the feasible range. In our prototype, we simply placed the main and auxiliary cameras tightly against each other. In the future, by using a beam splitter, the camera baseline can be further shortened, breaking through the physical size limitations of industrial cameras.
    

In summary, utilizing the proposed paraxial binocular SL system enables the achievement of unambiguous 3D reconstruction within the specified depth range, using only a single set of high-frequency fringe patterns. As a result, we can employ the cyclic projection strategy~\cite{zhong2013fast} for dynamic 3D scanning.

\section{Binomial Self-compensation for Motion Error }
In this paper, we cyclically project a group of high-frequency fringe patterns with a phase shift of $\pi/2$. The 3D point corresponding to a camera pixel $\left(x^c,y^c\right)$ is the intersection point of the sight line and the object surface. The intersection point moves along with the sight line on the circumstance that measures moving objects. Assuming the object is of low-frequency texture, the $i$~th captured frame can be expressed as
\begin{equation}\label{EQ:CapturedImage}
    {I_i\left(x^c,y^c\right)} = A^c + B^c\cos \left( {{\phi _0} - i\frac{\pi}{2}  + {x_i}} \right),
\end{equation}
where $A^c$ is the background intensity, $B^c$ is the modulation, $x_i$ represents the unknown phase offset induced by the motion from the datum frame to $i$-th frame, note that the datum frame can be arbitrarily chosen, and the true phase value at $i$~th frame is ${\phi _i} = {\phi _0} - i\frac{\pi}{2} $. We choose a $\pi/2$ phase shifting as it only transforms $\sin(\cdot)$ and $\cos(\cdot)$ terms into each other without inducing extra phase offset.

We notice that $x_i$ and its high order difference exactly correspond to the concepts of displacement, velocity, and acceleration, et. al, in kinematics, and can be expressed as
\begin{equation}\label{EQ:Difference}
\left\{ \begin{split}
{\Delta ^{\left( K \right)}}{x_i} = {\Delta ^{\left( {K - 1} \right)}}{x_{i + 1}} - {\Delta ^{\left( {K - 1} \right)}}{x_i}\\
{\Delta ^{\left( K \right)}}{x_i} = \sum\limits_{k = 0}^K {{{( - 1)}^k}} \left( \begin{split}
K\\
k
\end{split} \right){x_{i + K - k}}
\end{split} \right.,
\end{equation}
where we define ${\Delta ^{\left( 0 \right)}}{x_i}=x_i$. Eq.~(\ref{EQ:Difference}) is the key to motion error compensation, and we will use it later.

Utilizing a cyclic projection strategy~\cite{zhong2013fast} for high speed 3D scanning, a three or four-step phase shifting algorithm is typically employed to calculate the motion-affected phase as:
\begin{equation}\label{EQ:ErrorPhase}
    \tilde \phi _i^3 = {\tan ^{ - 1}}\left( {\frac{{\tilde S_i^3}}{{\tilde C_i^3}}} \right)~\text{and}~\tilde \phi _i^4 = {\tan ^{ - 1}}\left( {\frac{{\tilde S_i^4}}{{\tilde C_i^4}}} \right),
\end{equation}
where
\begin{equation}
        \left\{ \begin{split}
    {S_i^3} &= 2{I_{i + 1}} - {I_i} - {I_{i + 2}}\\
    {C_i^3} &= {I_i} - {I_{i + 2}}
    \end{split} \right.~\text{and}~
    \left\{ \begin{split}
    {S_i^4} &= {I_{i+1}} - {I_{i + 3}}\\
    {C_i^4} &= {I_i} - {I_{i + 2}}
    \end{split} \right..
\end{equation}
Please note that the "three-step phase shifting" discussed in this paper differs slightly from the traditional approach, as the phase shifting is $\pi/2$ instead of the traditional $2\pi/3$. We substitute Eq.~(\ref{EQ:CapturedImage}) into Eq.~(\ref{EQ:ErrorPhase}) to evaluate the motion error, for a small phase-shift error $x_i$, $\sin(x_i)\approx x_i$ and $\cos(x_i)\approx 1$, thus we have 
\begin{equation}\label{EQ:MotionError3}
\begin{split}
\epsilon_3 \left( {{\phi _i}} \right) &= {{\tilde \phi }_i^3} - {\phi _i}\\
 &\approx {\tan ^{ - 1}}\left( {\frac{{\left[ {1 + 0.5\left( {{x_i} - {x_{i + 2}}} \right)} \right]\sin \left( {{\phi _i}} \right) + \cos \left( {{\phi _i}} \right){x_{i + 1}}}}{{\cos \left( {{\phi _i}} \right) - 0.5\left( {{x_i} + {x_{i + 2}}} \right)\sin \left( {{\phi _i}} \right)}}} \right) - {\tan ^{ - 1}}\left( {\frac{{\sin \left( {{\phi _i}} \right)}}{{\cos \left( {{\phi _i}} \right)}}} \right)\\
 &\approx \frac{{{x_i} + 2{x_{i + 1}} + {x_{i + 2}}}}{4} + \frac{{2{x_{i + 1}} - \left( {{x_i} + {x_{i + 2}}} \right)}}{4}\cos \left( {2{\phi _i}} \right) + \frac{{{x_i} - {x_{i + 2}}}}{4}\sin \left( {2{\phi _i}} \right)\\
 & = \delta _i^3 + \frac{{2{x_{i + 1}} - \left( {{x_i} + {x_{i + 2}}} \right)}}{4}\cos \left( {2{\phi _i}} \right) + \frac{{{x_i} - {x_{i + 2}}}}{4}\sin \left( {2{\phi _i}} \right),
\end{split}
\end{equation}
and
\begin{equation}\label{EQ:MotionError4}
\begin{split}
{\epsilon _4}\left( {{\phi _i}} \right) &= \tilde \phi _i^4 - {\phi _i}\\
 &\approx {\tan ^{ - 1}}\left( {\frac{{2\sin \left( {{\phi _i}} \right) + \left( {{x_{i + 1}} + {x_{i + 3}}} \right)\cos \left( {{\phi _i}} \right)}}{{2\cos \left( {{\phi _i}} \right) - \left( {{x_i} + {x_{i + 2}}} \right)\sin \left( {{\phi _i}} \right)}}} \right) - {\tan ^{ - 1}}\left( {\frac{{\sin \left( {{\phi _i}} \right)}}{{\cos \left( {{\phi _i}} \right)}}} \right)\\
 &\approx \frac{{{x_i} + {x_{i + 2}} + {x_{i + 1}} + {x_{i + 3}}}}{4} + \frac{{{x_{i + 1}} - {x_i} + {x_{i + 3}} - {x_{i + 2}}}}{4}\cos \left( {2{\phi _i}} \right)\\
 & = \delta _i^4 + \frac{{{x_{i + 1}} - {x_i} + {x_{i + 3}} - {x_{i + 2}}}}{4}\cos \left( {2{\phi _i}} \right)
\end{split}
\end{equation}

According to Eq.~(\ref{EQ:MotionError3}) and~(\ref{EQ:MotionError4}), we discover that the motion error comprises a DC component and a harmonic component with twice the frequency of the wrapped phase. The DC component, manifesting as the lag of the point clouds due to object motion, is independent of the phase value, and thus is irrelevant to the ripple-shaped error. Consequently, the crux of motion error compensation lies in eliminating the harmonic.

\begin{figure}[t!]
    \centering
    \includegraphics[width=0.95\linewidth]{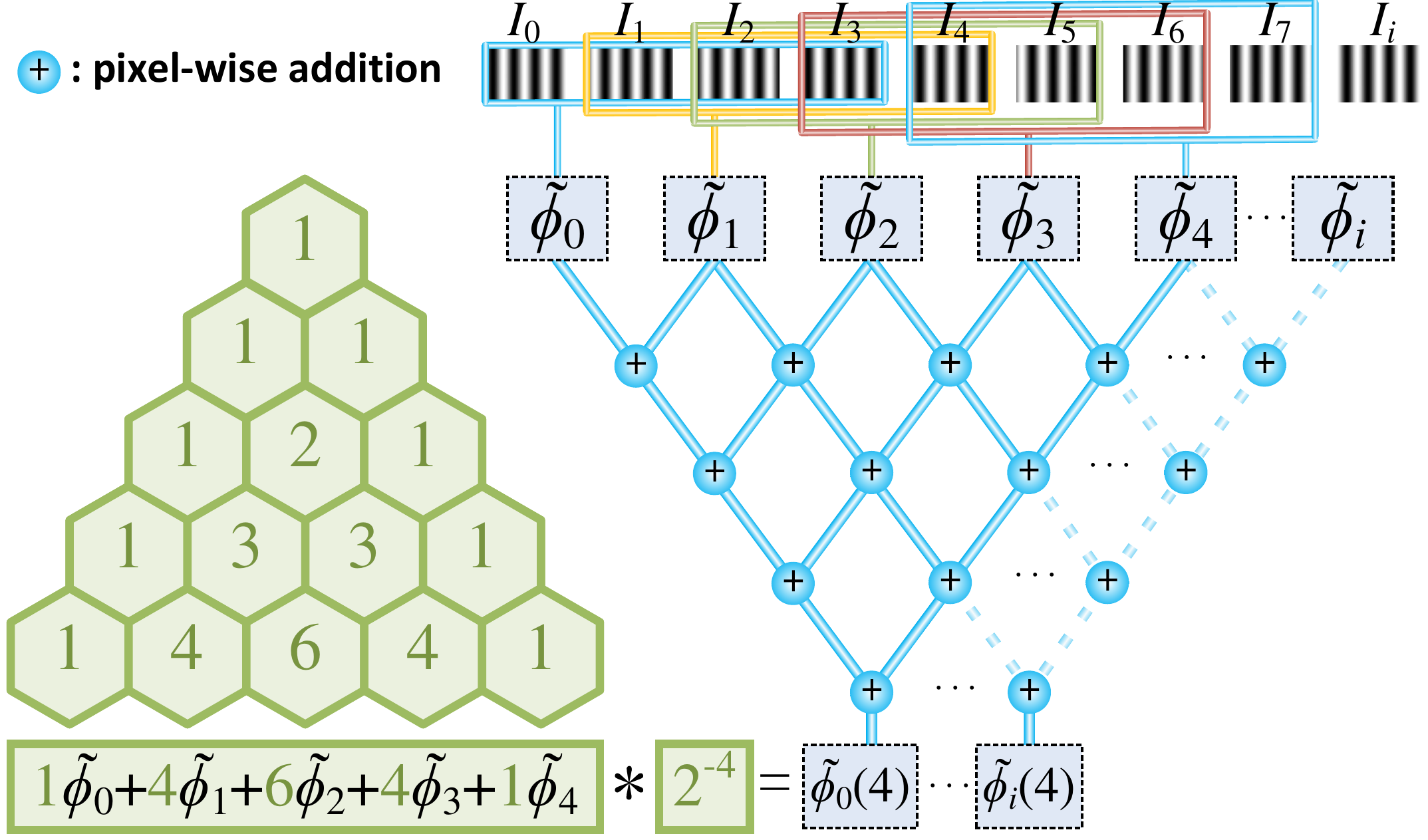}
    \caption{Binomial self-compensation for motion error.}
    \label{FIG:FigYanghui}
\end{figure}


Analyzing two successive phase frames $\tilde \phi _i$ and $\tilde \phi _{i+1}$ in a discrete time series, we discover that the coefficients of trigonometric terms exhibit opposite signs owing to a phase shift of $\pi/2$. Consequently, summing each pair of successive phase frames yields higher-order difference terms in the coefficients of harmonics. Repeating this process with the resultant phase frames, as depicted in Fig.~\ref{FIG:FigYanghui}, naturally evokes the concept of Yang Hui's triangle.

Let's consider calculating the weighted sum of $K+1$ successive phase frames in a time series as shown in Fig.~\ref{FIG:FigYanghui}, using the $(K+1)$-th row of Yang Hui's triangle, also as known as $K$-th order binomial coefficients
\begin{equation}\label{EQ:YangHuiSum}
\tilde \phi _i\left( K \right) = {2^{ - K}}\sum\limits_{k = 0}^K {\left( \begin{split}
K\\
k
\end{split} \right)\tilde \phi _{i + K - k}}\left( 0 \right),
\end{equation}
where $\tilde \phi _i\left( 0 \right) $ is computed by Eq.~(\ref{EQ:ErrorPhase}). We substitute Eq.~(\ref{EQ:Difference}) and (\ref{EQ:MotionError3}) into Eq.~(\ref{EQ:YangHuiSum}) to have
\begin{equation}
\begin{split}\label{EQ:YangHuiSum3Step}
    \tilde \phi _i^3\left( K \right) &= \bar \phi _i^3\left( K \right) - {2^{ - \left(K+2\right)}}{\Delta ^{\left( {K + 2} \right)}}{x_{i + 2}}\cos \left( {2{\phi _i}} \right) \\&+ {2^{ - \left(K+2\right)}}\left( {{\Delta ^{\left( {K + 1} \right)}}{x_{i + 2}} + {\Delta ^{\left( {K + 1} \right)}}{x_{i + 1}}} \right)\sin \left( {2{\phi _i}} \right)
\end{split},
\end{equation}
where
\begin{equation}
\bar \phi _i^3\left( K \right) = {2^{ - \left(K+2\right)}}\sum\limits_{k = 0}^K {\left[ {\left( \begin{split}
K\\
k
\end{split} \right)\left( {{\phi _{i + K - k}} + \delta _{i + K - k}^3} \right)} \right]}.
\end{equation}
Similarly, for four-step phase shifting we have 
\begin{equation}\label{EQ:YangHuiSum4Step}
    \tilde \phi _i^4\left( {K } \right) = \bar \phi _{i}^4\left( {K } \right) + {2^{ - \left(K+2\right)}}\left( {{\Delta ^{\left( {K + 1} \right)}}{x_i} + {\Delta ^{\left( {K + 1} \right)}}{x_{i + 2}}} \right)\cos \left( {2{\phi _i}} \right),
\end{equation}
where
\begin{equation}
    \bar \phi _{i}^4\left( {K } \right) = {2^{ - \left(K+2\right)}}\sum\limits_{k = 0}^K {\left[ {\left( \begin{array}{c}
    K\\
    k
    \end{array} \right)\left( {{\phi _{i + K - k}} + \delta _{i + K - k}^4} \right)} \right]}.
\end{equation}

\begin{figure}[t!] 
    \centering
    \subfloat[]
    {
        \includegraphics[width=0.47\linewidth]{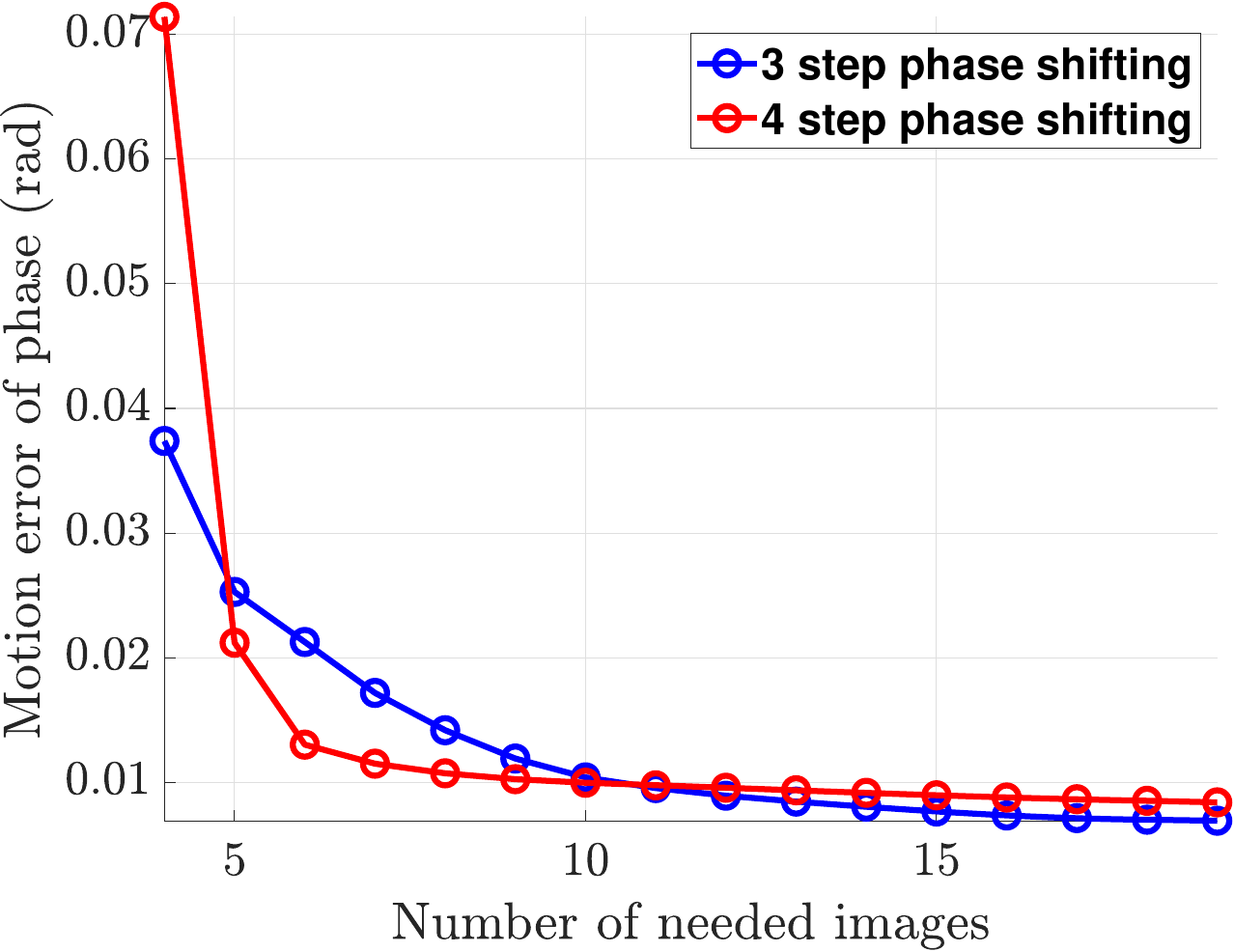}        
    }
    \subfloat[]
    {
        \includegraphics[width=0.5\linewidth]{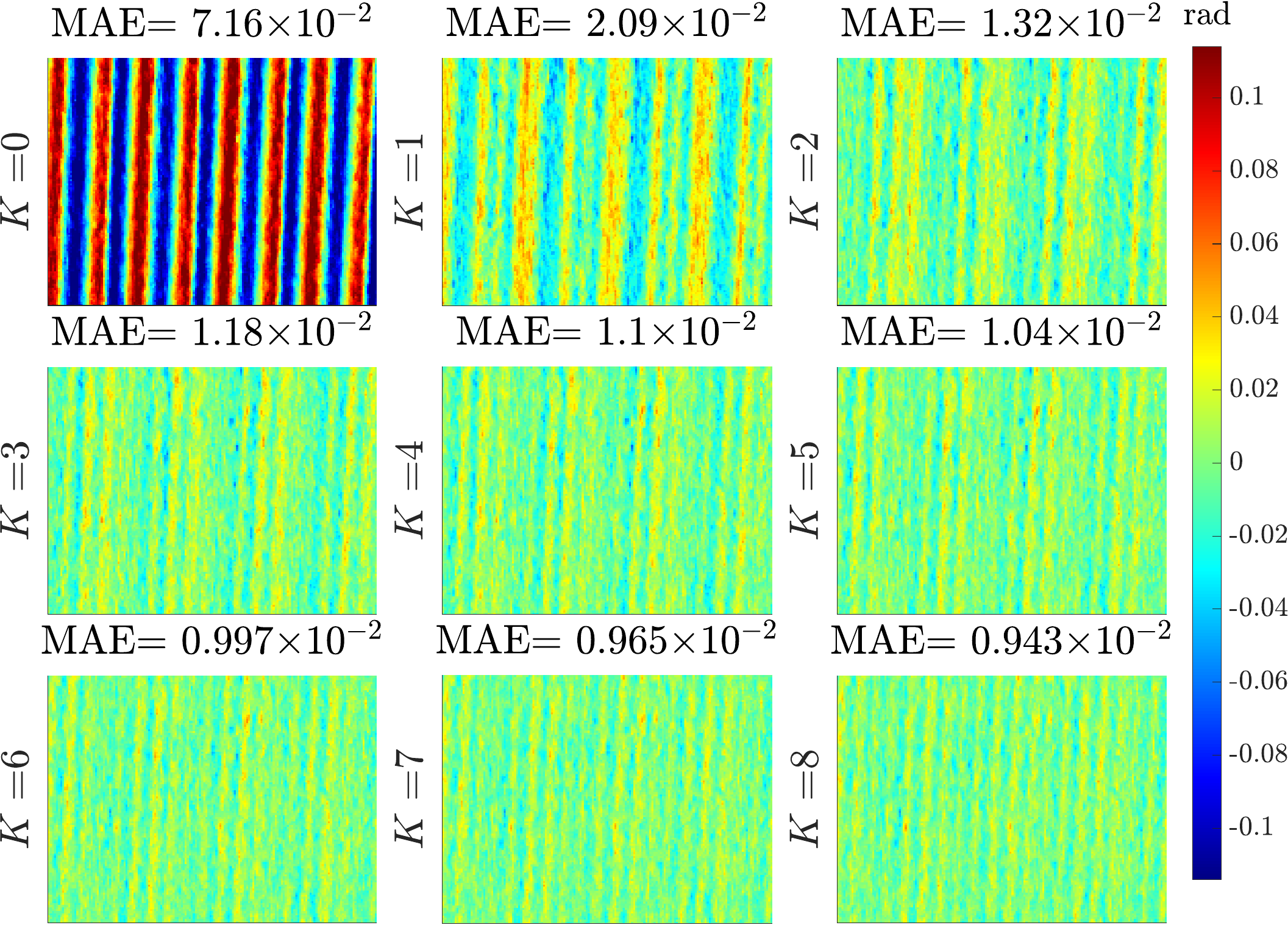}
    }
    \caption{ Motion error converges as $K$ increases: (a) three-step VS four-step phase shifting, and (b) visualization of motion error in four-step phase shifting, the mean absolute error of phase exponentially diminishes as $K$ increases.} 
    \label{FIG:FigPhaseErrorTok}
\end{figure}
Based on Eq.~(\ref{EQ:YangHuiSum3Step}) and (\ref{EQ:YangHuiSum4Step}), it is clear that increasing $K$ produces two main effects: first, the factor $2^{-(K+2)}$ exponentially decreases the harmonic amplitudes; second, higher-order difference terms (approach zero as $K$ increases) occur in the harmonic amplitudes. Both effects significantly diminish the motion error. 

We measure a flat plate moving towards the camera using our binomial self-compensation (BSC) for motion error, the results visualized in Fig.~\ref{FIG:FigPhaseErrorTok}(a) show that the ripples induced by motion in the self-compensated phase $\tilde \phi _i^4\left( K\right)$ reduces significantly with the increment of $K$. $K=0$ represents the raw phase that is affected by motion error. The residual error is caused by the nonlinearity of the binary defocus technique as shown in Fig.~\ref{FIG:FigPhaseErrorTok}(b). At least four frames of images are needed in BSC, and using eight images achieves a trade-off between accuracy and efficiency in practice. Note that the binomial order $K$ can be flexibly chosen according to the accuracy and time efficiency requirements.

In practice, we decompose Eq.~(\ref{EQ:YangHuiSum3Step}) into a pyramid-like procedure of adding up the successive two frames layer by layer, as shown in Fig.~\ref{FIG:FigYanghui}. However,
due to the ambiguity of the wrapped phase, phase order error occurs around the phase jumping area. To address this issue, we define the $\oplus$ operator as
\begin{equation}
{\phi _a} \oplus {\phi _b}: = \left\{ \begin{array}{l}
0.5\left( {{\phi _a} + {\phi _b}} \right),\left| {{\phi _a} - {\phi _b}} \right| > \pi \\
0.5\left( {{\phi _a} + {\phi _b}} \right) + \pi ,\left| {{\phi _a} - {\phi _b}} \right| < \pi 
\end{array} \right..
\end{equation}
This operation automatically compensates for the phase order error as long as the phase difference is greater than $\pi$. The motion-error-free phase $\tilde \phi_i\left( K\right)$ can be obtained by employing our binomial self-compensation (BSC), which is summarized as Algorithm~\ref{ALG:AlgCalibration}, where "$\bmod$" represents modular operation. Note that we have corrected the constant phase shifting before adding up two phase frames $\tilde \phi_i(0)$ and $\tilde \phi_{i+1}(0)$. The MATLAB code for implementing the algorithm are given in the github repository.

\section{Stereo Phase Unwrapping}
In the previous section, we obtained the motion-error-free high-frequency phase by conducting BSC. Subsequently, we employ a sum of absolute difference algorithm on the wrapped phase to retrieve the correspondence between the main and auxiliary cameras within the disparity range obtained by pre-set depth range. Further, the projector space coordinates $(x^p,y^p)$ can be computed by directly re-projecting the binocular 3D coordinate onto the projector space. Finally, we can have the phase order and compute the unwrapped phase.
\begin{algorithm}
    \small
    \caption{BSC for motion error} 
    \label{ALG:AlgCalibration}
    \KwIn
    {							
        \\Captured image sequence: $\left\{I_0,...,I_{T-1}\right\}$.
        \\Binomial order: $K$.
        \\Phase-shifting step: $N=3$~or~$4$.
    }
    \KwOut
    {
        \\Motion-error-free phase sequence: $\left\{\bar\phi_{0}(K),...,\bar\phi_{T-K-N+1}(K)\right\}$.			
    }			
    \For{ $t \leftarrow 0$ \KwTo $T-N$ }	
    {		
        $\tilde\phi_{t}(0) \leftarrow \left\{\begin{array}{c}
        \bmod \left[ {{{\tan }^{ - 1}}\left( \frac{2I_{t+1}-I_{t}-I_{t+2}}{I_t-I_{t+2}} \right)+t\frac{\pi }{2},2\pi } \right], N=3\\
        \bmod \left[ {{{\tan }^{ - 1}}\left( \frac{I_{t+1}-I_{t+3}}{I_t-I_{t+2}} \right)+t\frac{\pi }{2},2\pi } \right], N=4
        \end{array}\right.$
    }    
    \For{ $k \leftarrow 0$ \KwTo $K-1$ }	
    {		
        \For{ $i \leftarrow 0$ \KwTo $T-k-N$ }	
        {
            $\tilde\phi_{i}(k+1)\leftarrow \tilde\phi_{i}(k) \oplus \tilde\phi_{i+1}(k) $ 
        }
    }
\end{algorithm}

\section{Experimental Evaluations}
\begin{figure}[t!]
    \centering
    \includegraphics[width=0.7\linewidth]{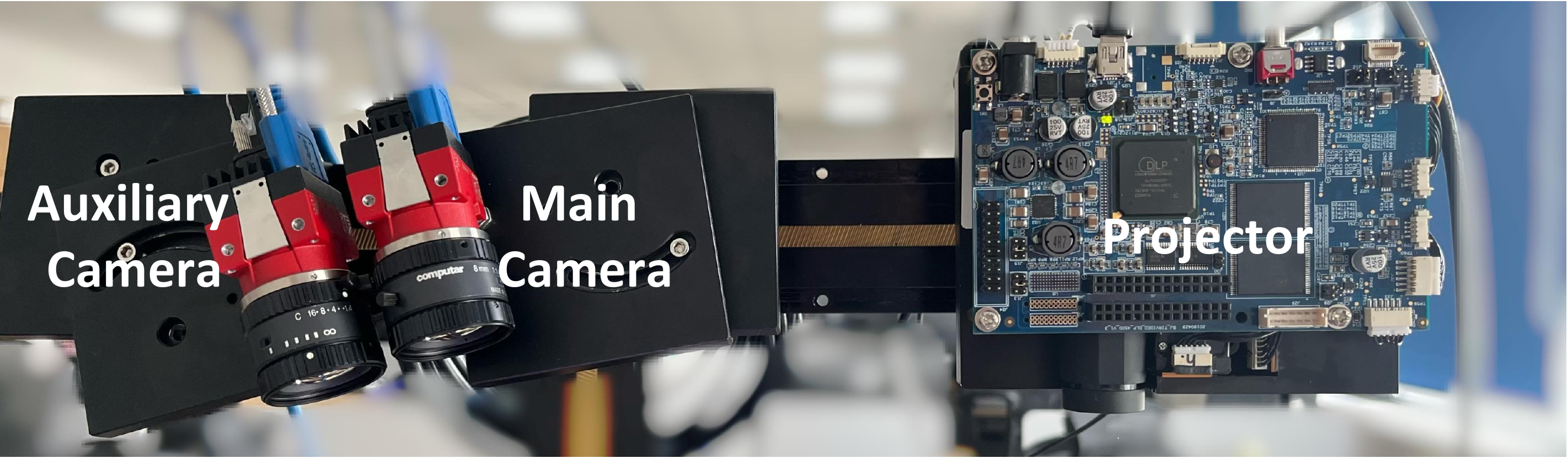}
    \caption{Experimental setup.}
    \label{FIG:FigExperimentSystem}
\end{figure}

The proposed algorithm is implemented on an Intel(R) Core(TM) i7-13790F @ 2.10 GHz with 32 GB RAM and written in MATLAB. Our experimental system~(Fig.~\ref{FIG:FigExperimentSystem}) consists of two AVT 1800U-120c cameras that have $1280\times960$ resolution, equipped with 8~mm Computar lenses, and a TI DLP4500 projector with a resolution of $912\times1140$. We cropped the cameras' field of view to $640\times480$ to have a 90~fps image acquisition speed. The two cameras are synchronized by the trigger signal from the projector.
In order to comprehensively demonstrate the superiority of our BSC, we conducted four groups of experiments on dynamic objects for evaluating: 1) absolute accuracy of 3D reconstruction, 2) temporal resolution of 3D imaging, 3) robustness on depth discontinuous scenes, and 4) generality on different dynamic objects.

\subsection{Absolute Accuracy of 3D Reconstruction}
\begin{figure}[h!] 
    \centering
    \includegraphics[width=0.7\linewidth]{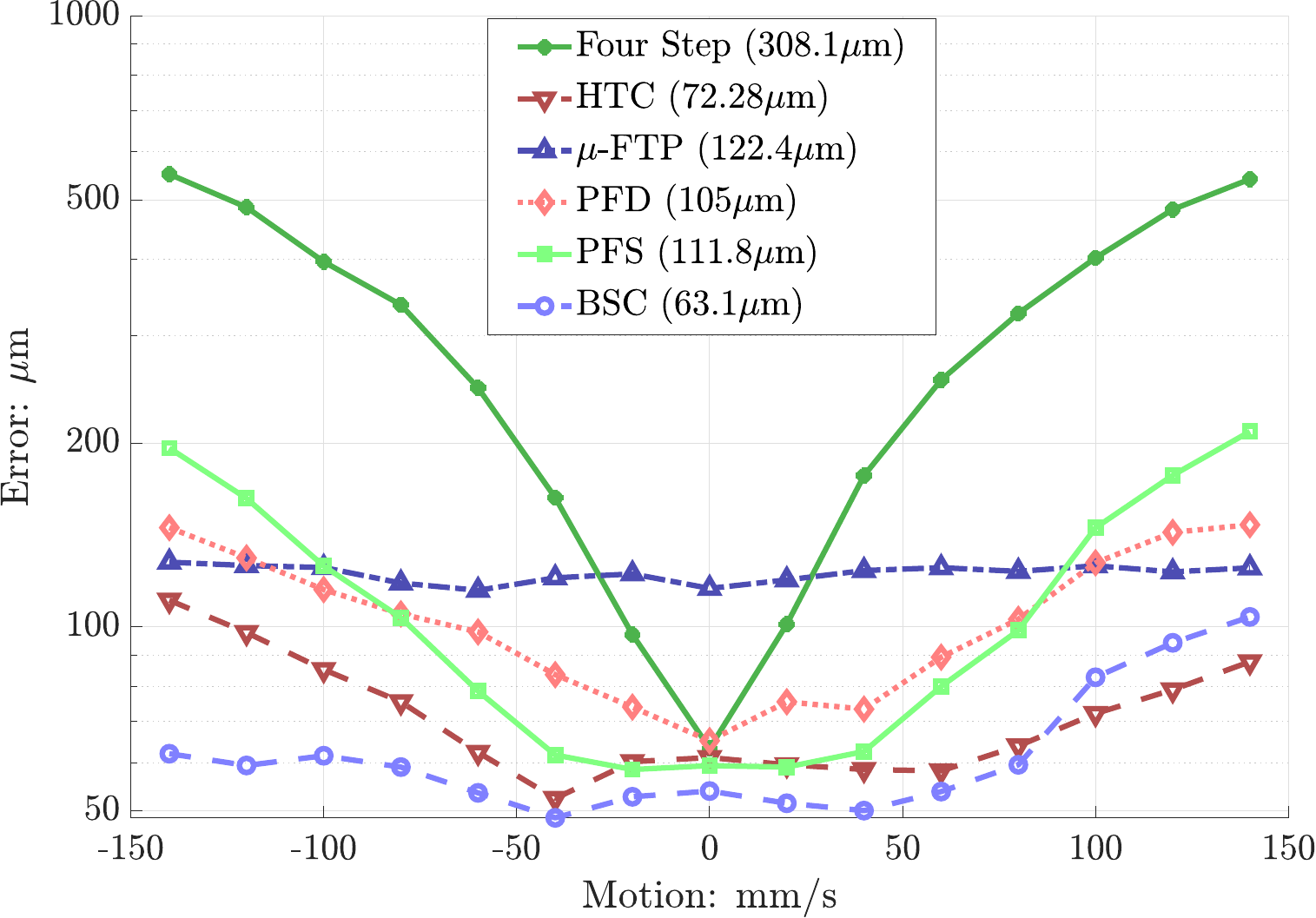}
    \caption{ Measurement error VS motion speed of a periodically waving plate by traditional four-step phase shifting, HTC~\cite{wang2018motion}, $\mu$-FTP~\cite{zuo2018micro}, PFD~\cite{liu2019real}, PFS~\cite{guo2021real}, and our BSC, the value in parentheses represents the mean error. Note the y-axis is in logarithmic scale.} 
    \label{FIG:FigExperiment1b}
\end{figure}

\begin{figure}[h!] 
    \centering
    \includegraphics[width=0.95\linewidth]{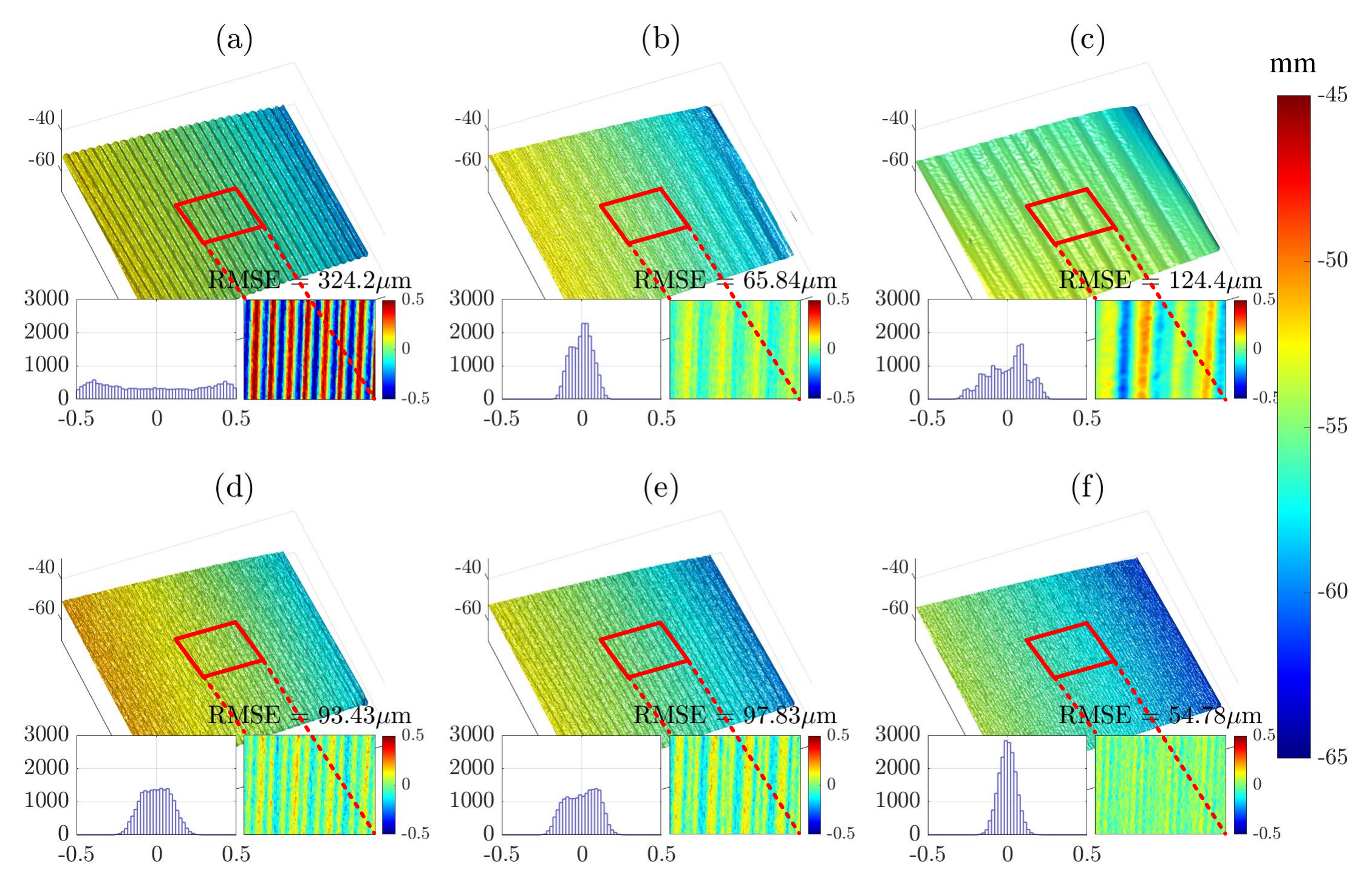}
    \caption{ Measurement results of a moving flat plane by (a) traditional four-step phase shifting, (b) HTC~\cite{wang2018motion}, (c)$\mu$-FTP~\cite{zuo2018micro}, (d) PFD~\cite{liu2019real}, (e) PFS~\cite{guo2021real}, and (f) our BSC.} 
    \label{FIG:FigExperiment1}
\end{figure}

We measured a periodically waving plate at a distance of 500~mm by using our BSC~(8 frames) and five representative methods for comparison: 1) traditional four-step PSP~(4 frames), 2) Hilbert transform compensation~(HTC, 4 frames)~\cite{wang2018motion}, 3) micro Fourier transform profilometry~($\mu$-FTP, 4 frames)~\cite{zuo2018micro}, 4) phase frame difference method~(PFD, 8 frames)~\cite{liu2019real}, 5) phase frame sum method~(PFS, 5 frmaes)~\cite{guo2021real}. The fringe wavelength was set to 24 pixels for the cyclic $\pi/2$ phase shifting patterns and was set to $[22,24,26]$ for $\mu$-FTP. The plate moves within a distance range of $[400,500]$~mm and a speed range of $[-150,150]$~mm/s, and our measurement lasted for 800 frames (about 8.9 seconds). We select three $160\times120$ rectangular windows in each depth map and fit the quadric surface as the ground truth, then we compute the root mean squared error (RMSE) within the selected regions. The trend of RMSE changing with motion speed is depicted in Fig.~\ref{FIG:FigExperiment1b}. The results show that our method exhibits the most satisfactory accuracy among existing methods when dealing with motion error.

Then we selected a moment when the speed of the plate was -88~mm/s, and depict the reconstructed point clouds in Fig.~\ref{FIG:FigExperiment1}, our BSC effectively reduces the motion error from 324.2~$\mu$m to 54.78~$\mu$m compared with traditional four-step phase shifting. By counting the histogram of errors, it can be observed that the error distribution of our BSC visually presents a normal distribution, which means Gaussian noise (induced by unstable ambient light, camera/projector flicker, quantization error, and sensor noise of the camera and projector) is the main component of error. In other words, the motion error is effectively suppressed by BSC. Meanwhile, the error distribution of other methods deviates from the normal distribution, indicating the existence of residual errors.
\subsection{Robustness on Depth Discontinuous Scenes}

\begin{figure}[t!] 
    \centering
    \includegraphics[width=0.98\linewidth]{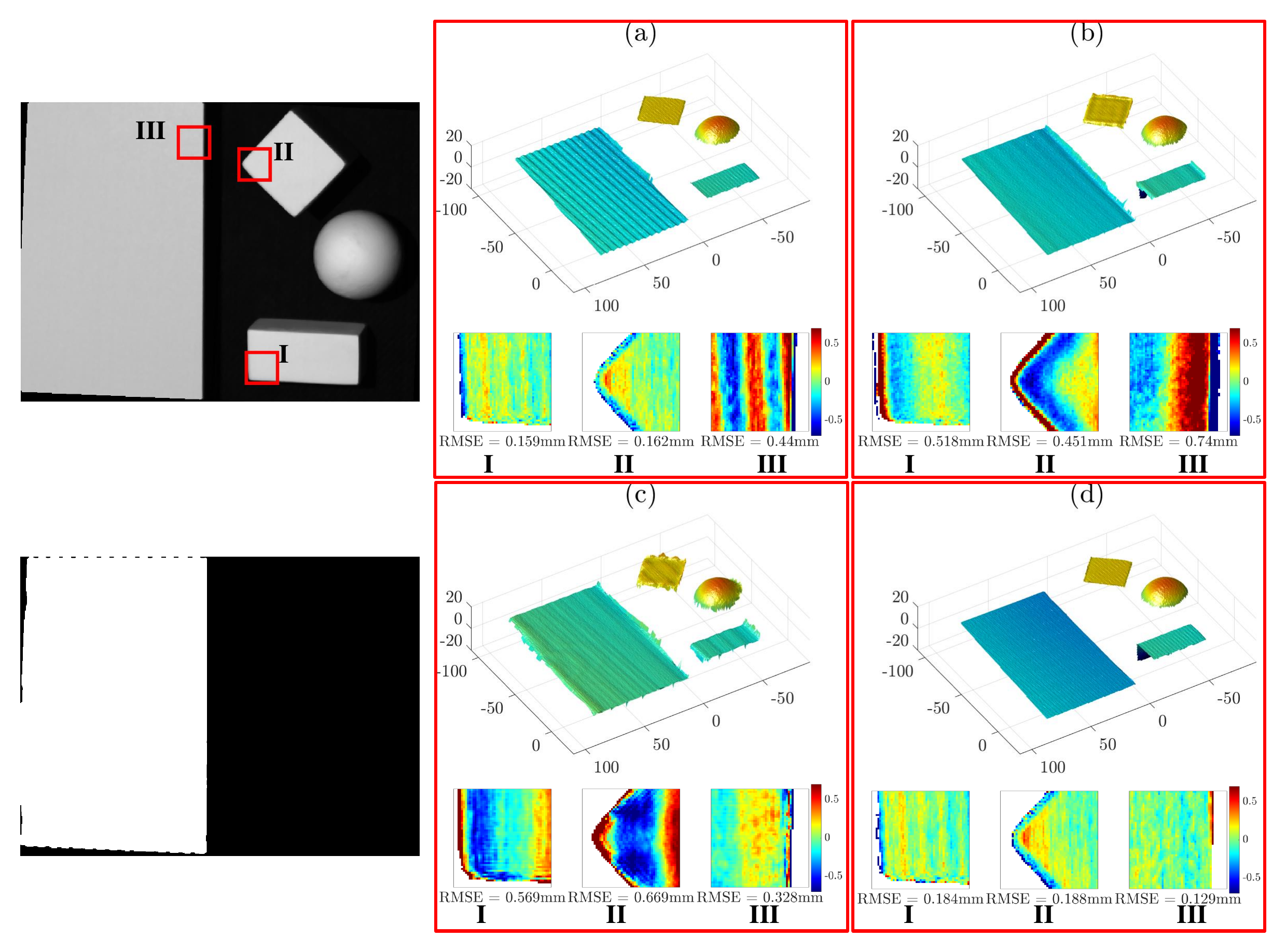}
    \caption{  Depth discontinuous scene (top left) and motion area labeled in white (bottom left). Reconstruction results and error at edges: (a) traditional four-step phase shifting, (b) HTC~\cite{wang2018motion}, (c) $\mu$-FTP~\cite{zuo2018micro}, and (d) our BSC.} 
    \label{FIG:FigExperiment2}
\end{figure}

PSP's pixel-wise characteristic is one of its most crucial merits. Our BSC maintains this property, thereby exhibiting robustness to depth discontinuous scenes. We set up a scene with three static gypsum geometries and one moving plate. Then we employed our BSC, $\mu$-FTP, and HTC. The results are shown in Fig.~\ref{FIG:FigExperiment2}. It can be seen that the two non-pixel-wise methods, i.e., $\mu$-FTP and HTC, exhibit visible deformation to the naked eye in the point cloud around the depth jumping areas. We selected three rectangular windows and quantitatively evaluate the RMSE, the results show that the RMSE of non-pixel-wise methods is significantly greater than our BSC. We can conclude that our BSC with pixel-wise property is more robust compared with non-pixel-wise methods when dealing with depth discontinuous scenes.
\subsection{Temporal Resolution of 3D Imaging}
\begin{figure}[t!] 
    \centering
    \includegraphics[width=0.98\linewidth]{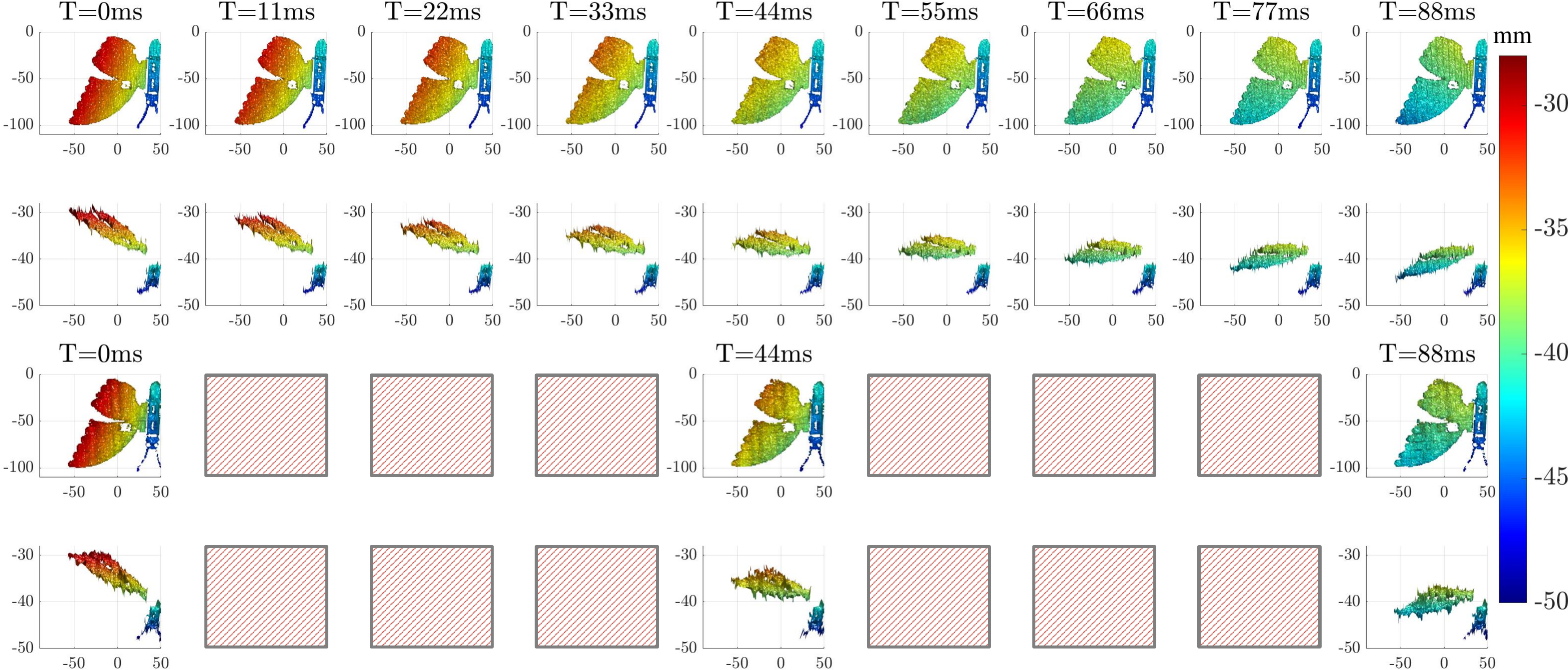}
    \caption{  The measurement result of a wooden butterfly with wings flapping, during T=0~ms $\sim$ 88~ms, \textbf{Row 1 and 2}: our BSC; \textbf{Row 3 and 4}: $\mu$-FTP~\cite{zuo2018micro}.} 
    \label{FIG:FigExperiment3}
\end{figure}
Our BSC employs cyclic $\pi/2$ phase-shifting fringe patterns, any $K+N$ successive captured images can be used for reconstructing one frame of 3D point clouds. Therefore, our resulting frame rate of 3D point clouds is the same as the camera frame rate, thereby achieving a high temporal resolution of the measurement. That is the frame-wise loopable property we mentioned in the introduction. We successively conducted BSC and $\mu$-FTP to measure a wooden butterfly with wings flapping cyclically, and then manually aligned the timeline, the results are shown in Fig.~\ref{FIG:FigExperiment3}. We noticed that our method can achieve 3D reconstruction with a frame rate of 90~fps~(the same as the camera frame rate), while $\mu$-FTP reconstructing at 30~fps. Thus, our method is capable of distinguishing tinier motions.
\subsection{Generality on Different Dynamic Objects}
We captured 400 image frames (about 4 seconds) employing our pipeline on four different dynamic objects, and the results are shown in Fig.~\ref{FIG:FigExperiment4}. The complete motion process video can be found in the supplementary materials. We observed that BSC can effectively eliminate the severe ripple errors caused by object motion.
\begin{figure}[t!]
    \centering
    \includegraphics[width=0.98\linewidth]{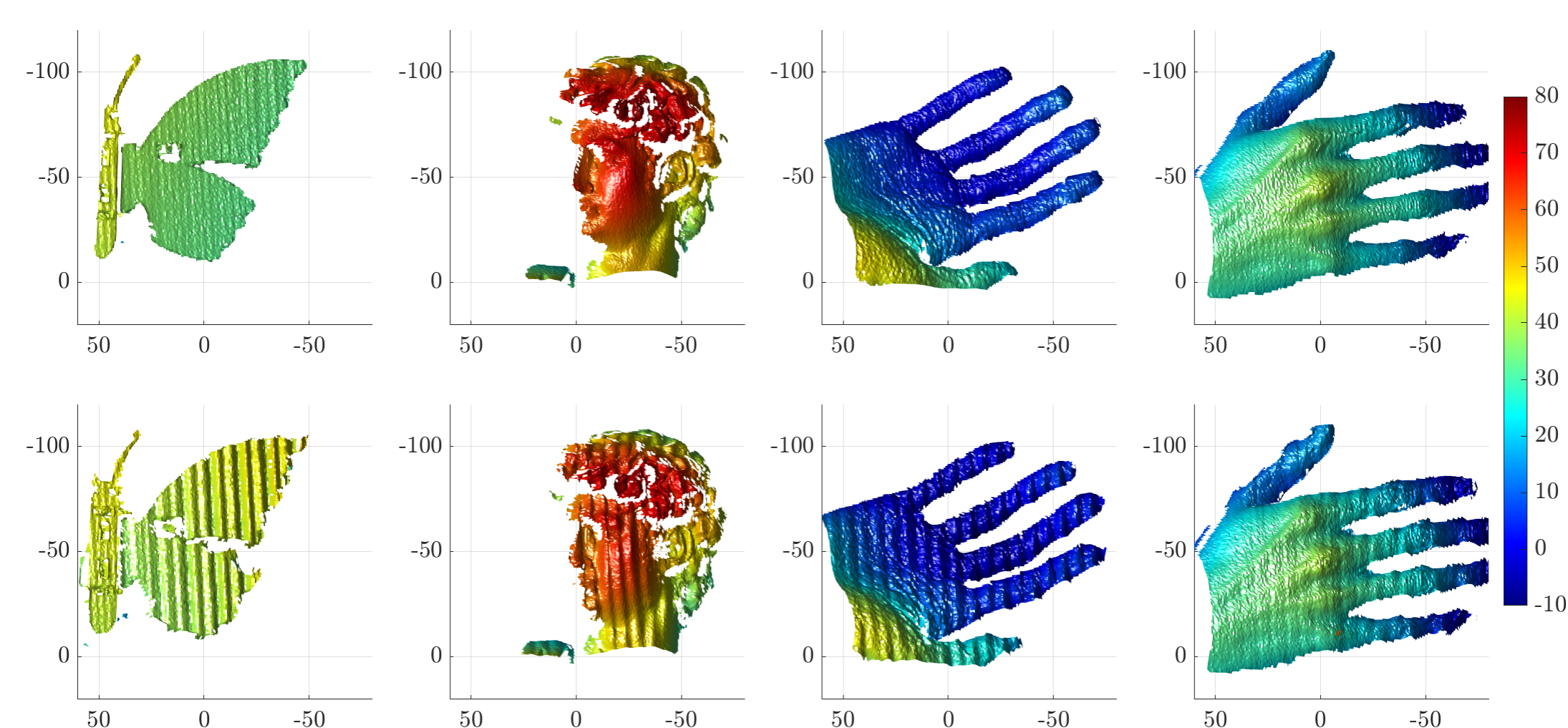}
    \caption{The measurement results of four different dynamic objects. \textbf{Column 1}: wooden butterfly model with wings flapping, \textbf{Column 2}: moving statue, \textbf{Column 3}: waving hand, \textbf{Column 4}: hand showing scissors, stone, and cloth gestures. \textbf{Top to bottom}: our BSC and traditional four-step PSP.}
    \label{FIG:FigExperiment4}
\end{figure}

\section{Limitations and Future Outlook}
\begin{figure}[t!]
    \centering
    \includegraphics[width=0.5\linewidth]{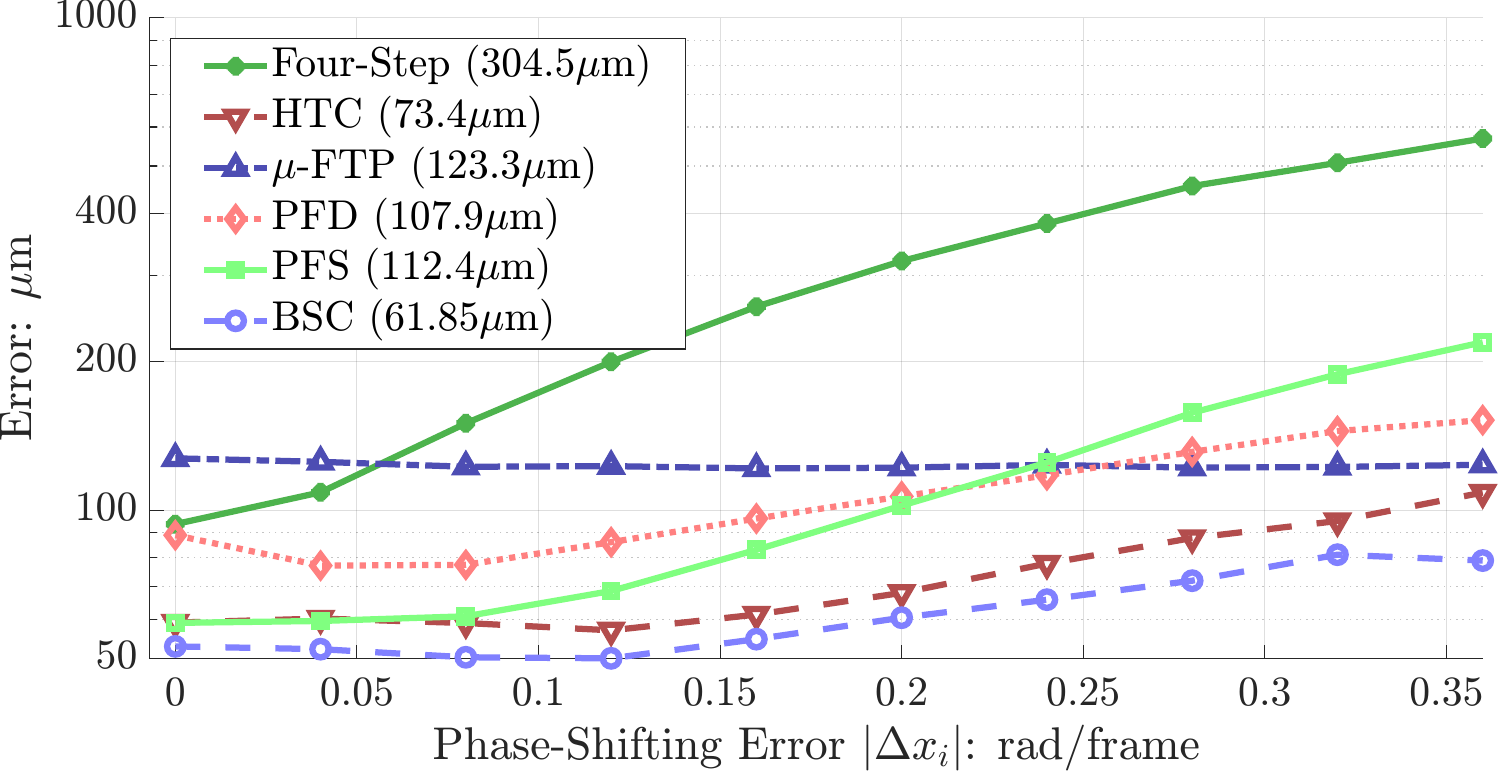}
    \caption{3D error VS inter-frame phase-shifting error $|\Delta x_i|$.}
    \label{FIG:FigPhaseDifferenceVSError}
\end{figure}
We presented a pixel-wise and frame-wise loopable binomial self-compensation algorithm to effectively and flexibly eliminate motion error in four-step PSP, without any intermediate variables assisted. 

\textbf{High-frequency textures}: our BSC assumes low-frequency textures like most motion compensation algorithms, it is robust against moderate contrast textures. However, when dealing with objects that have high contrast textures, our method induces artifacts similar to those seen with typical phase-shifting methods. 

\textbf{Phase unwrapping error}: we adopt a common block matching algorithm for stereo correspondence, inducing obivious phase unwrapping error when measuring complex surfaces. In the future, we will conduct a hierarchical adjustment for phase outliers to optimize the accuracy of the point cloud.

\textbf{Applicable speed range}: Our BSC is predicated on the assumption that the phase-shift error $x_i$ is small. Consequently, as shown in Fig.~\ref{FIG:FigPhaseDifferenceVSError}, the residual motion error arises as $|\Delta x_i|$ increases. Properly reducing the fringe frequency could decrease $|\Delta x_i|$, alleviating the performance loss of BSC for high-speed objects.
%
%
\bibliographystyle{splncs04}
\bibliography{manuscript}
\end{CJK}
\end{document}